\documentclass[twoside]{article}

\usepackage[preprint]{aistats2026}
\usepackage[round]{natbib}

\usepackage{amsmath}
\usepackage{graphicx}
\usepackage[hidelinks]{hyperref}
\usepackage{doi}
\usepackage{amssymb}
\usepackage{mdframed}
\usepackage{amsthm}
\usepackage{cleveref}
\usepackage{array}
\usepackage{float}

\begin{document}

%

%

\twocolumn[

	\aistatstitle{Variance-Aware Prior-Based Tree Policies for Monte Carlo Tree Search}

	\aistatsauthor{ Maximilian Weichart }

	\aistatsaddress{ University of Tübingen } ]

\begin{abstract}
	Monte Carlo Tree Search (MCTS) has profoundly influenced reinforcement learning (RL) by
	integrating planning and learning in tasks requiring long-horizon reasoning, exemplified by the
	AlphaZero family of algorithms. Central to MCTS is the search strategy, governed by a tree
	policy based on an upper confidence bound (UCB) applied to trees (UCT). A key factor in the
	success of AlphaZero is the introduction of a prior term in the \textit{UCB1}-based tree policy
	\textit{PUCT}, which improves exploration efficiency and thus accelerates training. While many
	alternative UCBs with stronger theoretical guarantees than \textit{UCB1} exist, extending them
	to prior-based UCTs has been challenging, since \textit{PUCT} was derived empirically rather
	than from first principles. Recent work retrospectively justified \textit{PUCT} by framing MCTS
	as a regularized policy optimization (RPO) problem. Building on this perspective, we introduce
	\emph{Inverse-RPO}, a general methodology that systematically derives prior-based UCTs from a
	broad class of prior-free UCBs. Applying this method to the variance-aware \textit{UCB-V},
	we obtain two new
	prior-based tree policies that incorporate variance estimates into the search. Experiments
	indicate that these variance-aware prior-based UCTs outperform \textit{PUCT} across multiple
	benchmarks without incurring additional computational cost. We also provide an extension of the
	\texttt{mctx} library supporting variance-aware UCTs, showing that the required code changes are
	minimal and intended to facilitate further research on principled prior-based UCTs. Code:
	\url{https://github.com/Max-We/inverse-rpo}.
\end{abstract}

\section{Introduction}

The combination of reinforcement learning (RL) with Monte Carlo Tree Search (MCTS) has led to major
advances in artificial intelligence. Starting with AlphaGo \citep{silverMasteringGameGo2016}, and
subsequently generalized by AlphaZero \citep{silverGeneralReinforcementLearning2018} and MuZero
\citep{schrittwieserMasteringAtariGo2020}, this line of work has achieved superhuman performance
across domains requiring long-horizon reasoning and complex decision-making. These results
underscore the power of integrating learning with search-based planning, and they motivate ongoing
efforts to develop more efficient and broadly applicable variants of MCTS and AlphaZero-style
methods.

A central component of MCTS is the tree policy, which balances exploration and exploitation to
minimize regret. Before AlphaZero, such policies were derived from upper confidence bounds (UCBs)
such as \textit{UCB1} \citep{auerFinitetimeAnalysisMultiarmed2002}, giving rise to the well-studied family of UCT algorithms, which apply UCBs to
tree search. Over time, many variants beyond \textit{UCB1}---including \textit{UCB-V},
\textit{Bayesian UCT}, and \textit{UCB1-Uniform/Power} \citep{audibertExplorationExploitationTradeoff2009,tesauroBayesianInferenceMonteCarlo2012,asaiExtremeValueMonte2024}---have been explored and shown to have a
significant effect on the MCTS performance.  With the AlphaZero family of algorithms, \textit{UCB1}
was extended by incorporating a prior term estimated by a neural network, yielding \textit{PUCT}.
This prior-based extension of \textit{UCB1} greatly improved search efficiency in both small and
large action spaces \citep{wuMiniZeroComparativeAnalysis2023} and has since become the de facto
standard tree policy.  However, extending this prior-based approach to other UCBs has proven
difficult. While the authors claim that \textit{PUCT} is a variant of
\textit{PUCB}~\citep{rosinMultiarmedBanditsEpisode2011}, which in itself is an extension of
\textit{UCB1} with contextual information, a complete proof was never presented. Indeed, the
concrete form of \textit{PUCT} deviates from \textit{UCB1} and \textit{PUCB} by introducing a
heuristic decay of the exploration term, and it is generally assumed to have been derived
empirically rather than from formal guarantees\footnote{See the discussion by
	\cite{grillMonteCarloTreeSearch2020} or the historical context in a
	\href{https://groups.google.com/g/computer-go-archive/c/K9XHb64JSqU}{Google Groups thread}.}. We
hypothesize that the extension of other UCBs to prior-based UCTs in the context of MCTS, although
promising in theory, has been underexplored for that reason.

\begin{table}[h]
	\centering
	\caption{Four prior-based UCT rules arranged by base UCB (columns) and heuristic form (rows).
		The heuristic form of the UCTs is described in Section~\ref{sec:mcts}. Our contributions are
		marked with *.}
	\label{tab:uct-grid}
	\vspace{0.5em} 
	\begin{tabular}{lcc}
		\hline
		                        & \textbf{UCB1} & \textbf{UCB-V} \\
		\hline
		\textit{canonical form} & UCT-P         & UCT-V-P*       \\
		\textit{heuristic form} & PUCT          & PUCT-V*        \\
		\hline
	\end{tabular}
\end{table}

Recent work has reinterpreted MCTS as regularized policy optimization (RPO), showing that
\textit{PUCT} can be viewed as tracking the solution to a specific RPO. Our key insight is that this
perspective not only provides an understanding for the form of prior-based UCBs in hindsight, such
as previously described for \textit{PUCT} \citep{grillMonteCarloTreeSearch2020}, but also the theoretical
foundation needed to systematically derive a broad class of prior-based UCTs directly from
prior-free UCB bonuses by expressing them as an RPO.  Building on this insight, we continue to study prior-based UCTs beyond
\textit{PUCT} by extending other, potentially stronger, UCB-based policies with prior terms. More
concretely, we make the following key contributions:

\textbf{Inverse-RPO.} We introduce \emph{Inverse-RPO}, a principled, step-by-step method that
transforms a decomposable UCB bonus into its prior-based counterpart. Unlike prior work that starts from an already
prior-based selector such as \textit{PUCT} \citep{grillMonteCarloTreeSearch2020}, our method derives
a prior-based selector systematically from its prior-free base form (e.g., \textit{UCB1}). While
prior work provides the formal framework linking MCTS and UCTs to RPO
\citep{grillMonteCarloTreeSearch2020}, we rearrange and slightly extend this approach into an
easy-to-follow methodology, enabling researchers to apply it directly to their UCB of choice in
future work.

\textbf{Variance-Aware Prior-Based UCTs.} To explore prior-based UCTs beyond \textit{PUCT}, we instantiate
\emph{Inverse-RPO} on the variance-aware \textit{UCB-V} to obtain two prior-based tree policies (see
Table~\ref{tab:uct-grid}): (i) \textit{UCT-V-P}, a principled RPO-derived variant; and (ii)
\textit{PUCT-V}, a heuristic analogue aligned with the practical form of \textit{PUCT}. As
experimental baselines, we compare these derived tree-policies against \textit{PUCT} (the de facto
choice in the AlphaZero family of algorithms), while also benchmarking against \textit{UCT-P}
\citep{grillMonteCarloTreeSearch2020}, which can be viewed as a prior-based \textit{UCB1} without
the heuristic alterations introduced with \textit{PUCT}.

\textbf{Empirical Validation and Implementation.} Across a range of benchmark domains, we show that
our variance-aware prior-based \textit{UCT-V-P} and \textit{PUCT-V} consistently match or outperform
\textit{UCT-P} and \textit{PUCT} respectively, indicating that the benefits of replacing
\textit{UCB1} with stronger UCBs such as \textit{UCB-V} in MCTS extend naturally to the prior-based
MCTS as in the AlphaZero family of algorithms. We further propose an efficient implementation
strategy for variance-aware MCTS, demonstrating that the derived \textit{UCT-V-P} and
\textit{PUCT-V} can be deployed in practice as easily as the commonly used \textit{PUCT} and at no
extra computational overhead.

\section{Preliminaries}

Before presenting our methodology, we briefly review the key background concepts and notation needed
throughout the paper. We begin with Monte Carlo Tree Search (MCTS) and its standard UCT formulation,
followed by the regularized policy optimization (RPO) perspective that provides the foundation for
our derivations.

\subsection{Monte Carlo Tree Search}
\label{sec:mcts}

Monte Carlo Tree Search (MCTS) is a widely used planning algorithm that incrementally builds a
search tree through repeated simulations (see Appendix~\ref{app:mcts}). During search, a tree policy
based on an upper confidence bound (UCB) balances exploration and exploitation
\citep{kocsisBanditBasedMonteCarlo2006}\footnote{Notation: \emph{(1)} We use \emph{UCB/UCT} in
	upright font as generic descriptors for the family of upper confidence bound rules (UCT denotes a
	UCB applied to trees). \emph{(2)} Concrete algorithms/instantiations are written in italics (e.g.,
	\textit{UCB-V}, \textit{PUCT}). \emph{(3)} The canonical Hoeffding-based forms are written
	\textit{UCB1}/\textit{UCT1} to distinguish them from the generic descriptors in (1). A suffix ``-P''
	indicates a prior-based extension (e.g., \textit{UCT-P}, \textit{PUCT-V}).}. When \textit{UCB1} is
applied to trees, this yields the classical upper confidence bound for trees (\textit{UCT1})
\citep{kocsisBanditBasedMonteCarlo2006}:

\begin{equation}
	\pi_{\mathrm{UCT1}}  :=  \arg\max_a
	\left[ q_a + c \cdot
		\sqrt{\frac{\log N}{1+n_a}} \right].
	\label{eq:uct}
\end{equation}

Here $q_a$ is the empirical action value, $n_a$ its visit count, and $N=\sum_b n_b$ the total visits
at the node. \textit{UCT1} is provably optimal in the sense that it achieves the correct
exploration-exploitation trade-off and converges to the optimal policy as the number of visits
grows. Throughout this work, we add $1$ to the visit count $n_a$, without loss of generality, to
avoid division by zero and to simplify the subsequent analysis.

The action selection rule used in AlphaZero, commonly referred to as \textit{PUCT}
\citep{silverAlphaZeroChessMastering2017}, was introduced later. It augments \textit{UCB1} with
the policy prior $\pi_\theta(a)$, which is being approximated by a neural network.

\begin{equation}
	\label{eq:puct}
	\pi_{\mathrm{PUCT}}  :=  \arg\max_a
	\left[ q_a + c \cdot \pi_\theta(a) \cdot
		\frac{\sqrt{N}}{1 + n_a} \right].
\end{equation}

\textbf{PUCT Heuristic Exploration Decay.}
Besides the prior term, \textit{PUCT} \eqref{eq:puct} departs from the principled \textit{UCB1} rule
by adopting a different exploration bonus that scales only with the square root of the total visit
count $N$, rather than with $\sqrt{\log N}$.  Formally, this amounts to replacing the exploration
term
\[
	\sqrt{\tfrac{\log N}{1+n_a}}
\]
in \textit{UCT1} ~\eqref{eq:uct} with
\[
	\frac{\sqrt{N}}{1+n_a}.
\]

Later, \cite{grillMonteCarloTreeSearch2020} proposed a principled variant,
\textit{UCT-P}, which, similar to \textit{PUCT} extends \textit{UCB1} by incorporating the policy
prior, but without the heuristic exploration decay.

\begin{equation}
	\pi_{\mathrm{UCT\mbox{-}P}}  :=  \arg\max_a
	\left[ q_a + c \cdot
		\sqrt{ \pi_\theta(a) \cdot \frac{\log N}{1+n_a}} \right].
	\label{eq:uct-p}
\end{equation}

By formalizing MCTS as a regularized policy optimization (RPO) problem, they showed that \textit{UCT-P}
directly expresses an RPO and that even \textit{PUCT} can be cast within this framework—thus providing a
theoretical justification in hindsight for its heuristic form.

\subsection{Regularized Policy Optimization}

Many machine-learning problems have been expressed as convex optimization problems
\citep{bubeckConvexOptimizationAlgorithms2015}, such as Support Vector Machines (SVMs)
\citep{scholkopfLearningKernelsSupport2001} or Trust Region Policy Optimization (TRPO)
\citep{schulmanTrustRegionPolicy2017}. Equivalently, reinforcement learning (RL) can be interpreted
as a convex optimization problem by expressing it as RPO
\begin{equation}
	\pi_{\theta'}  :=  \arg\max_{\mathbf{y} \in \mathcal{S}}
	\Big[ \mathbf{q}^\top \mathbf{y} - \mathcal{R}(\mathbf{y}, \pi_\theta) \Big],
\end{equation}

where $\mathbf{y}$ is a distribution over actions, $\mathbf{q}$ the corresponding $q$-values, and
$\mathcal{R}: \mathcal{S}^2 \rightarrow \mathbb{R}$ a divergence-based convex regularizer that keeps
$\mathbf{y}$ close to the prior policy $\pi_\theta$
\citep{neuUnifiedViewEntropyregularized2017,geistTheoryRegularizedMarkov2019,grillMonteCarloTreeSearch2020}.

\citet{grillMonteCarloTreeSearch2020} proved that MCTS with \textit{UCT1} \eqref{eq:uct} corresponds to the
solution of an RPO with the Hellinger distance:

\begin{equation}
	\begin{aligned}
		\label{eq:uct-p-rpo}
		\bar{\pi}_{\mathrm{UCT\mbox{-}P}}  :=
		\arg\max_{\mathbf{y} \in \mathcal{S}}
		\Big[\mathbf{q}^\top \mathbf{y} - \lambda^{\mathrm{UCT\mbox{-}P}}_{N} D_{\mathrm{H}}(\pi_\theta,\mathbf{y})\Big], \\
		\lambda^{\mathrm{UCT\mbox{-}P}}_{N}  :=
		c \cdot \sqrt{\frac{\log N}{|\mathcal{A}| + N}}.
	\end{aligned}
\end{equation}

where $\mathcal{A}$ denotes the action set and $\mathcal{S}$ is the $|\mathcal{A}|$-dimensional probability
simplex.

Similarly, they showed that \textit{PUCT} \eqref{eq:puct} expresses the solution to an RPO with the
reverse-KL distance:

\begin{equation}
	\begin{aligned}
		\label{eq:puct-rpo}
		\bar\pi_{\mathrm{PUCT}}  :=  \arg\max_{\mathbf{y} \in \mathcal{S}}
		\Big[ \mathbf{q}^\top \mathbf{y} - \lambda^{\mathrm{PUCT}}_N \, D_{\mathrm{KL}}(\pi_\theta, \mathbf{y}) \Big], \\
		\lambda^{\mathrm{PUCT}}_N  :=  c \cdot
		\frac{\sqrt{N}}{|\mathcal{A}| + N}.
	\end{aligned}
\end{equation}

From this RPO perspective, the \textit{UCT-P} \eqref{eq:uct-p} and \textit{PUCT} \eqref{eq:puct}
can be recovered by considering the optimal action of the RPOs and evaluating the \emph{marginal
	one-step gain} when selecting action $a$. Following prior work, we keep the notation
$\frac{\partial}{\partial n_a}$; operationally, this denotes the change along the coupled MCTS
update in which both $n_a$ and the total count $N=\sum_b n_b$ increase by one.
\begin{equation}
	a^\star_{\mathrm{UCT\mbox{-}P}} = \arg\max_a
	\left[ \frac{\partial}{\partial n_a}
		\big( \mathbf{q}^\top \hat{\pi} - \lambda^{\mathrm{UCT\mbox{-}P}}_N \, D_{\mathrm{H}}(\pi_\theta, \hat{\pi}) \big) \right]
\end{equation}

\begin{equation}
	a^\star_{\mathrm{PUCT}} = \arg\max_a
	\left[ \frac{\partial}{\partial n_a}
		\big( \mathbf{q}^\top \hat{\pi} - \lambda^{\mathrm{PUCT}}_N \, D_{\mathrm{KL}}(\pi_\theta, \hat{\pi}) \big) \right]
\end{equation}

\section{Deriving Prior-Based UCTs: Inverse RPO Pipeline}
\label{sec:inverse-rpo}

While previous work has established the existence of an RPO formulation for prior-based UCTs, a
clear methodology to derive such a prior-based UCT starting from a canonical prior-free UCB bonus has been
missing.  Our first contribution is therefore \emph{methodological}: we propose an
\emph{Inverse-RPO} pipeline, summarized in Figure~\ref{fig:inverse-rpo}, which provides a systematic
procedure to derive prior-based UCTs from a broad class of prior-free UCB bonuses and offers
researchers a principled framework to follow:

\begin{figure}
	\centering
	\includegraphics[width=0.5\textwidth]{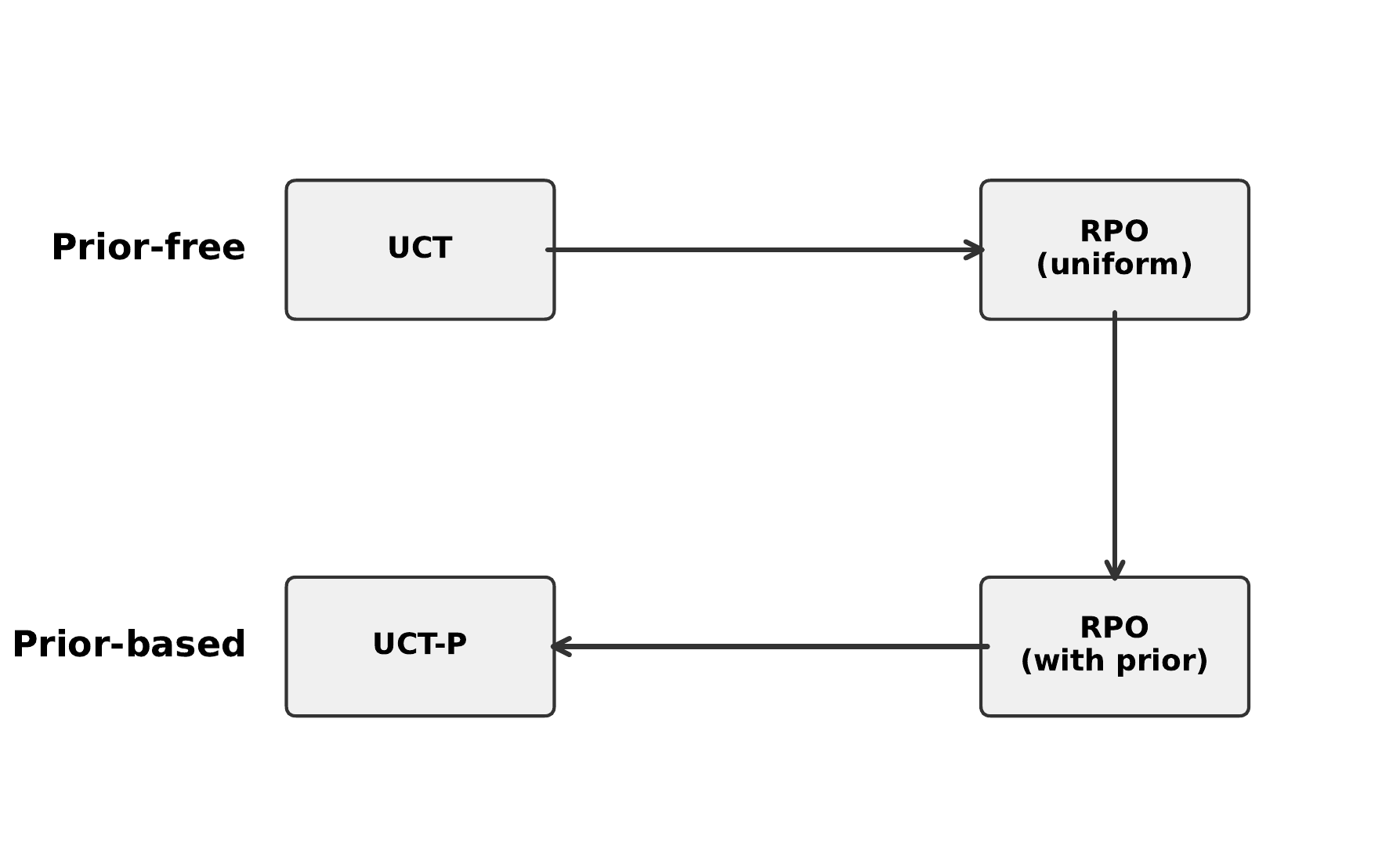}
	\caption{By casting the MCTS tree policy as the solution to an RPO objective, the prior becomes
		an explicit design term, yielding a principled prior-based UCT selection rule. This perspective
		resolves the otherwise opaque step from prior-free UCT to prior-based UCT and motivates our
		\emph{Inverse-RPO} methodology.}
	\label{fig:inverse-rpo}
\end{figure}

\noindent We summarize the \emph{Inverse-RPO} pipeline below and, at each step, give a compact worked
example showing how \textit{UCT-P} follows from \textit{UCT1}.

\paragraph{Step 1: Decompose the exploration bonus into action-independent and action-dependent factors.}
Rewrite the exploration bonus in terms of the empirical visit distribution $\hat\pi(a)$, separating
an action-independent scaling factor $\Phi(N)$ from an action-dependent weighting function
$h(\hat\pi(a))$.
\par\smallskip
\noindent\textit{Example (UCT1).} Define
\begin{equation}
	\label{eq:selector}
	\hat\pi(a)=\frac{1+n_a}{|\mathcal A|+N}, \qquad N=\sum_b n_b.
\end{equation}
\noindent\emph{Substitute} $\hat\pi(a)$ to obtain
\begin{equation}
	c\,\sqrt{\tfrac{\log N}{1+n_a}}
	=
	\underbrace{\tfrac{c\,\sqrt{\log N}}{\sqrt{|\mathcal A|+N}}}_{\Phi(N)}
	\cdot
	\underbrace{\tfrac{1}{\sqrt{\hat\pi(a)}}}_{h(\hat\pi(a))},
	\qquad
	h(x)=x^{-1/2}.
\end{equation}

\paragraph{Step 2: Define a (separable) regularizer generated by $f$.}
Choose a convex function $f$ whose derivative matches the negative action-dependent weighting
function, ${f}'(x)=-h(x)$. This defines a prior-free (separable) regularizer generated by $f$ and
hence a prior-free RPO.
\par\smallskip
\noindent\textit{Example (UCT1).} With $h(x)=x^{-1/2}$, one may take $f(x)=2(1-\sqrt{x})$ since
$f'(x)=-x^{-1/2}=-h(x)$.
The corresponding (separable) regularizer is
\begin{equation}
	\sum_a 2(1-\sqrt{y_a}).
\end{equation}

\paragraph{Step 3: Lift the (separable) regularizer to an $f$-divergence.}
Replace the (separable) regularizer generated by $f$ with the corresponding $f$-divergence
$D_f(\pi_\theta,y)$ to obtain an explicit prior-based RPO.
\par\smallskip
\noindent\textit{Example (UCT1).} In general, a prior-free regularizer of the form
$\sum_a f(y_a)$ corresponds to the $f$-divergence $\sum_a \pi_\theta(a)
	f\!\big(y_a/\pi_\theta(a)\big)$. For \textit{UCT1}, this yields the Hellinger-type $f$-divergence
\begin{equation}
	D_{\mathrm{H}}(\pi_\theta,y) \;=\; \sum_{a} \pi_\theta(a)\, 2\!\left(1-\sqrt{\tfrac{y_a}{\pi_\theta(a)}}\right).
\end{equation}

\paragraph{Step 4: Recover the selector.}
Take the marginal gain with respect to $n_a$ to obtain the prior-based UCT rule. By the definition
in Step 2, this corresponds to replacing the action-dependent weighting function
$h(\hat\pi(a))$ by $h\!\big(\hat\pi(a)/\pi_\theta(a)\big)$.
\par\smallskip
\noindent\textit{Example (UCT1).} The lift replaces the action-dependent weighting function
$h(\hat\pi(a))$ by
$h\!\big(\hat\pi(a)/\pi_\theta(a)\big)$, so the bonus becomes
\begin{equation}
	\Phi(N)\,h\!\left(\tfrac{\hat\pi(a)}{\pi_\theta(a)}\right)
	= c\,\sqrt{\pi_\theta(a)\cdot \tfrac{\log N}{1+n_a}},
\end{equation}
recovering \textit{UCT-P} (Eq.~\eqref{eq:uct-p}).

\noindent\emph{Recap.} \emph{Inverse-RPO} introduces priors by first expressing a prior-free
UCT bonus as the derivative of a (separable) regularizer, then lifting that regularizer to the
corresponding $f$-divergence with prior $\pi_\theta$. For \textit{UCT1} this yields \textit{UCT-P},
and for a uniform $\pi_\theta$, it reduces to the prior-free form (up to a constant factor).

\section{UCT-V-P and PUCT-V: Variance-Aware Prior-based UCTs}
\label{sec:puct-v}

Our aim is to go beyond \textit{UCB1}, studying alternative base UCBs with tighter confidence
bonuses and deriving their prior-based counterparts via the \emph{Inverse-RPO} pipeline. A natural
candidate is \textit{UCB-V}, which augments the exploration bonus with an empirical-variance term
and is obtained from a Bernstein-type concentration inequality (in contrast to the Hoeffding
inequality underlying \textit{UCB1}) \citep{audibertExplorationExploitationTradeoff2009}. Under the
same bounded-reward assumption, this yields variance-adaptive bonuses and correspondingly tighter
instance-dependent guarantees than \textit{UCB1}, without changing the problem setting. The
variance-aware UCB-V applied to MCTS
\citep{audibertExplorationExploitationTradeoff2009,wissowScaleAdaptiveBalancingExploration2024} is
\begin{equation}
	\label{eq:uct-v}
	\begin{aligned}
		S_a^{\mathrm{UCT\mbox{-}V}}(q,n,N)
		 & = q_a + B^{\mathrm{UCT\mbox{-}V}}(N,n_a,\hat\sigma_a^2),                                \\
		B^{\mathrm{UCT\mbox{-}V}}(N,n_a,\hat\sigma_a^2)
		 & :=  c_1 \,\hat\sigma_a\,\sqrt{\tfrac{\log N}{1+n_a}} \;+\; c_2 \,\tfrac{\log N}{1+n_a},
	\end{aligned}
\end{equation}
where $\hat\sigma_a$ is the empirical reward standard deviation for action $a$ consistent with
earlier notation. We set $c_1 = \sqrt{2}$ and $c_2 = 3$, so that the above expression is
algebraically identical to the definition of \cite{audibertExplorationExploitationTradeoff2009},
with the constants absorbed into $c_1$ and $c_2$.

Analogous to the \textit{PUCT} exploration-decay heuristic (see Section~\ref{sec:mcts}), we
introduce a heuristic variant, \textit{UCT-V-H}, which rewrites the exploration bonus as
shown in~\eqref{eq:uct-v-h}. This heuristic form is introduced to make the comparison
with \textit{PUCT} meaningful as a whole; without it, we could only compare against the
principled baseline \textit{UCT-P}.
\begin{equation}
	\label{eq:uct-v-h}
	B^{\mathrm{UCT\mbox{-}V{-}H}}(N,n_a,\hat\sigma_a^2)
	= c_1 \hat\sigma_a\,\tfrac{\sqrt{N}}{1+n_a} + c_2 \,\tfrac{\log N}{1+n_a}.
\end{equation}

We apply the \emph{Inverse-RPO} pipeline to obtain variance-aware, prior-based counterparts of
\textit{UCT-V} and its heuristic decay \textit{UCT-V-H}. Specifically, the pipeline yields (i)
UCT-style \emph{selection rules} that can be used as drop-in replacements for
\textit{PUCT}/\textit{UCT-P} during tree traversal and (ii) corresponding \emph{RPO objectives}
that mirror the selection rules in the optimization view of MCTS.

\begin{center}
	\fbox{\parbox{0.96\linewidth}{
			\textbf{Result: Variance-aware prior-based UCT selection rules.} \\

			\noindent\textbf{UCT-V-P:}
			\begin{equation}
				\label{eq:uct-v-p}
				\begin{aligned}
					S_a^{\mathrm{UCT\mbox{-}V\mbox{-}P}}(q,n,N)
					= q_a
					+ c_1 \cdot \hat\sigma_a \, \sqrt{ \pi_\theta(a)\, \tfrac{\log N}{1+n_a} } \\
					+ c_2 \cdot \pi_\theta(a) \, \tfrac{\log N}{1+n_a}.
				\end{aligned}
			\end{equation}

			\noindent\textbf{PUCT-V:}
			\begin{equation}
				\label{eq:puct-v}
				\begin{aligned}
					S_a^{\mathrm{PUCT\mbox{-}V}}(q,n,N)
					= q_a
					+ c_1 \cdot \pi_\theta(a)\, \hat\sigma_a \, \tfrac{\sqrt{N}}{1+n_a} \\
					+ c_2 \cdot \pi_\theta(a) \, \tfrac{\log N}{1+n_a}.
				\end{aligned}
			\end{equation}
			\emph{Derivations:} see Appendix~\ref{app:uctv-derivations}.
		}}
\end{center}

\noindent\emph{Notable elements (selectors).} (i) The prior enters the exploration bonus as
$\pi_\theta(a)$, reweighting both the variance and bias terms of \textit{UCB-V}. (ii) The placement
of the prior inside a square root for \textit{UCT-V-P} follows from the divergences used in the
Inverse-RPO lift (Hellinger vs.\ reverse-KL) and is specified in the next RPO objectives (other
box). (iii) For a uniform prior, both selectors reduce to their prior-free forms.

\begin{center}
	\fbox{\parbox{0.96\linewidth}{
			\textbf{Result: Variance-aware prior-based RPO targets.}\\[4pt]

			\noindent\textbf{UCT-V-P:}
			\begin{equation}
				\label{eq:L-uct-v-p}
				\begin{aligned}
					L_{\mathrm{UCT\mbox{-}V\mbox{-}P}}(y)
					 & = \mathbf{q}^\top y
					- \lambda_{N}^{\mathrm{UCT\mbox{-}V-1}}\, D_{\mathrm{H}}^{\hat\sigma}(\pi_\theta,y) \\
					 & \quad - \lambda_{N}^{\mathrm{UCT\mbox{-}V-2}}\, D_{\mathrm{KL}}(\pi_\theta,y).
				\end{aligned}
			\end{equation}
			\begin{equation}
				\begin{aligned}
					\lambda_{N}^{\mathrm{UCT\mbox{-}V-1}}
					= c_1 \frac{\sqrt{\log N}}{\sqrt{|\mathcal A|+N}},
					\lambda_{N}^{\mathrm{UCT\mbox{-}V-2}}
					= c_2 \frac{\log N}{|\mathcal A|+N}.
				\end{aligned}
			\end{equation}

			\noindent\textbf{PUCT-V:}
			\begin{equation}
				\label{eq:L-puct-v}
				\begin{aligned}
					L_{\mathrm{PUCT\mbox{-}V}}(y)
					 & = \mathbf{q}^\top y
					- \lambda_{N}^{\mathrm{UCT\mbox{-}V\mbox{-}H-1}}\, D_{\mathrm{KL}}^{\hat\sigma}(\pi_\theta,y) \\
					 & \quad - \lambda_{N}^{\mathrm{UCT\mbox{-}V\mbox{-}H-2}}\, D_{\mathrm{KL}}(\pi_\theta,y).
				\end{aligned}
			\end{equation}
			\begin{equation}
				\lambda_{N}^{\mathrm{UCT\mbox{-}V\mbox{-}H-1}}
				= c_1 \frac{\sqrt{N}}{|\mathcal A|+N},
				\lambda_{N}^{\mathrm{UCT\mbox{-}V\mbox{-}H-2}}
				= c_2 \frac{\log N}{|\mathcal A|+N}.
			\end{equation}

			\medskip
			\noindent\emph{Derivations:} see Appendix~\ref{app:uctv-derivations}.
		}}
\end{center}

\noindent\emph{Notable elements (RPO objectives).}
(i) In contrast to the \textit{UCT-P} \eqref{eq:uct-p-rpo} and \textit{PUCT} \eqref{eq:puct-rpo}
optimization targets, which use a \emph{single} regularizer term with one weight $\lambda_N$, our
variance-aware contributions use \emph{two} regularizer terms with distinct weights: a variance-term
weight $\lambda_N^{(1)}$ and a bias-term weight $\lambda_N^{(2)}$.  (ii) As a result of the
heuristic form of \textit{UCT-V-H} in line with \textit{PUCT}, the two variance-aware objectives are
identical in their \emph{second} regularizer term, and they differ only in the \emph{first}
regularizer and its weight.  (iii) The first regularizer in each objective is a
\emph{variance-weighted $f$-divergence term}: $D_{\mathrm{H}}^{\hat\sigma}(\pi_\theta,y)$ for \textit{UCT-V-P} and
$D_{\mathrm{KL}}^{\hat\sigma}(\pi_\theta,y)$ for \textit{PUCT-V}, where the per-action factors
$\hat\sigma_a$ are treated as data-dependent side information (not optimization variables in the RPO
over $y$). Formal definitions are given in Appendix~\ref{app:uctv-derivations}.

\section{Experiments}
\label{sec:experiments}

Our experimental aim is twofold: (i) to implement the new variance-aware UCT policies
\textit{PUCT-V} and \textit{UCT-V-P} introduced in Section~\ref{sec:puct-v}; and (ii) to evaluate their
performance relative to the classical prior-based baselines \textit{PUCT} and \textit{UCT-P}.
We first describe the implementation details of the variance-aware extensions before turning
to empirical comparisons.

\subsection{Variance-aware MCTS Implementation}

We provide a variance-aware MCTS implementation by extending the
\texttt{mctx}\footnote{\url{https://github.com/google-deepmind/mctx}} library
\citep{deepmind2020jax}. Enabling \textit{UCT-V}-style rules requires propagating both empirical
means and variances from a leaf to the root. To this end, we adopt Welford's online update (see
Algorithm~1), which is numerically stable and adds only a constant-time, constant-memory
augmentation to the standard mean backpropagation \citep{welfordNoteMethodCalculating1962}.
Concretely, each node stores $(n, \mu, \sigma^2)$ instead of $(n, \mu)$, where $n$ is the visit
count. The control flow and backward pass remain identical to standard mean backpropagation, with
the starred (${}^\star$) lines denoting the added variance-tracking updates. During the selection
phase, we also incorporate the proposed \textit{PUCT-V} and \textit{UCT-V-P} rules.

In the AlphaZero framework, a neural network is trained to approximate both the value function and
the empirical visit distribution produced by MCTS. For our purposes, no additional variance head is
required and the empirical variance from the tree search is sufficient.

\begin{samepage}
	\vspace{0.5em}
	\begin{mdframed}[nobreak=true]
		\noindent\textbf{Algorithm 1} Variance-aware single-node update\\
		\textbf{Input:} parent stats $(n,\mu,\sigma^2)$; discounted value $v \!=\! r + \gamma \cdot v_{\text{child}}$.\\[2pt]
		\noindent\textbf{Complexity:} each update requires $\mathcal{O}(1)$ arithmetic operations and $\mathcal{O}(1)$ memory.\\[4pt]
		\[
			\begin{aligned}
				n^{+}       & \leftarrow n + 1,                                                   \\
				\Delta      & \leftarrow v - \mu,                                                 \\
				\mu^{+}     & \leftarrow \mu + \tfrac{\Delta}{n^{+}},                             \\
				\Delta_2    & \leftarrow v - \mu^{+}                                   & {}^\star \\[3pt]
				\sigma^{2+} & \leftarrow \tfrac{n\,\sigma^2 + \Delta\,\Delta_2}{n^{+}} & {}^\star
			\end{aligned}
		\]
		\textbf{Return:} $(n^{+},\,\mu^{+},\,\sigma^{2+})$.
	\end{mdframed}
\end{samepage}

Overall, adapting MCTS to be variance-aware and to use the proposed
selection rules requires only three lines of code, excluding the additional variance field in the
data structures.

\subsection{Bandit Evaluation}

\noindent We first use a two-armed Bernoulli bandit to isolate how the prior term and the
empirical-variance term affect exploration in \textit{UCT-V-P} and \textit{PUCT-V}.

\begin{figure*}[!t]
	\centering
	\includegraphics[width=0.95\linewidth]{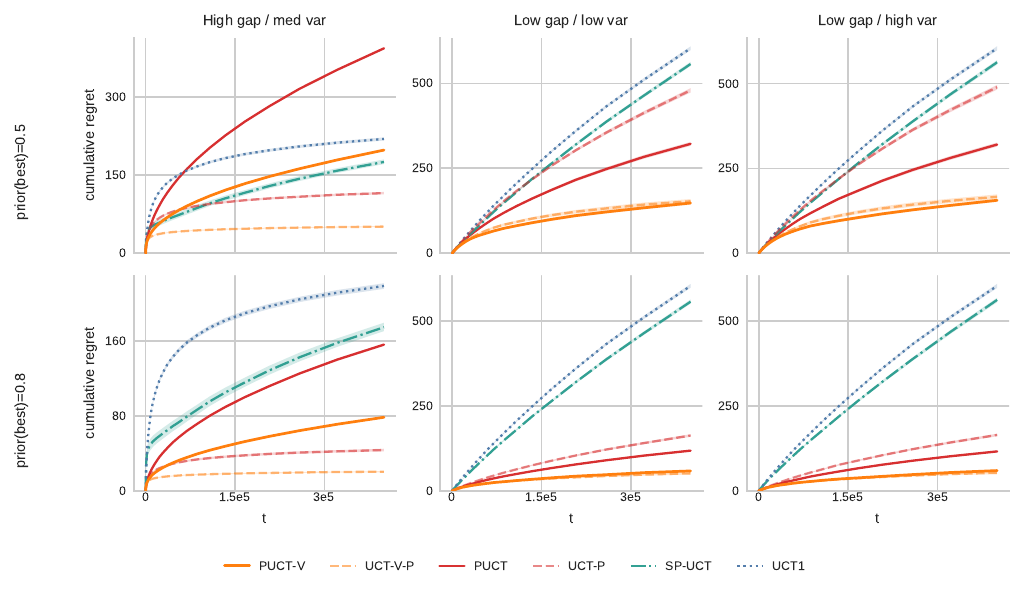}
	\caption{Two-armed Bernoulli bandit evaluation: cumulative regret over time. Columns correspond
		to the three bandit regimes (high gap / medium variance; low gap / low variance; low gap /
		high variance). Rows vary the quality of the action prior used by prior-based methods:
		\textbf{top:} an uninformative uniform prior ($\pi(\text{best})=0.5$); \textbf{bottom:} an
		informative prior ($\pi(\text{best})=0.8$).}
	\label{fig:bandit-regret}
\end{figure*}

\paragraph{Setup}
Following \citet{wolfFrameworkFairEvaluation2025a}, we consider three regimes
(Figure~\ref{fig:bandit-regret}): \textbf{(i)} high gap / medium variance ($\mu=(0.8,\,0.9)$),
\textbf{(ii)} low gap / low variance ($\mu=(0.895,\,0.900)$), and \textbf{(iii)} low gap / high
variance ($\mu=(0.890,\,0.895)$). Each method runs for $T=4\times 10^5$ pulls and we report cumulative
regret.

\paragraph{Selectors and priors}
We compare against \textit{UCT1}\footnote{Strictly speaking, this section evaluates bandits (so
	\emph{UCB}, not UCT). We keep the term \textit{UCT} for consistency with the rest of the paper:
	indicating ``$+1$'' in the denominator.} \citep{auerFinitetimeAnalysisMultiarmed2002} as the canonical
baseline; \textit{PUCT} and \textit{UCT-P} as prior-based, variance-unaware comparators; and
\textit{SP-UCT} (Stochastic Power UCT) \citep{damPowerMeanEstimation2024} as a strong prior-free
alternative base form. We use the default selector hyperparameters summarized in
Appendix~\ref{app:selector-hparams}. To test the interaction with the prior term, we evaluate an
uninformative uniform prior ($\pi(\text{best})=0.5$) and an informative prior
($\pi(\text{best})=0.8$).

\paragraph{Observations.}
Across all three regimes and both prior settings, the variance-aware selectors improve substantially
over their variance-unaware counterparts: \textit{PUCT-V} consistently reduces regret relative to
\textit{PUCT}, and \textit{UCT-V-P} improves over \textit{UCT-P}. The gap is most prominent in the
low-gap regimes, where variance-awareness stabilizes exploration and yields the lowest regret,
consistent with \citet{wolfFrameworkFairEvaluation2025a}. In the high-gap regime, the variance-aware
selectors remain among the best, but the ranking among variance-unaware baselines shifts, suggesting
that when the suboptimality gap is large the underlying UCT base form may dominate and variance
plays a secondary role. Finally, an informative prior substantially reduces regret for all
prior-based methods, highlighting the value of even moderately accurate priors, such as those
introduced into UCT-selectors via \emph{Inverse-RPO} (Section~\ref{sec:inverse-rpo}).

\subsection{Synthetic Tree Evaluation}

\noindent We now extend our analysis to trees and study how performance of the derived
variance-aware prior-based selectors varies with depth and branching factor.

\begin{figure*}[!t]
	\centering
	\includegraphics[width=0.95\linewidth]{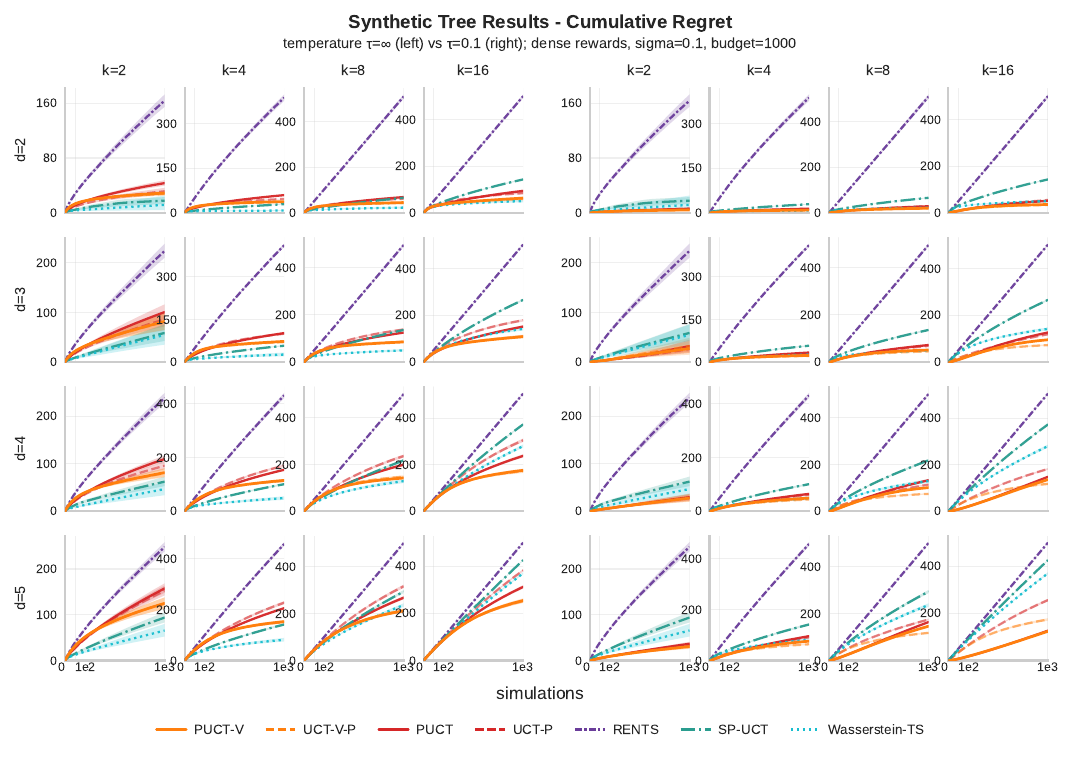}
	\caption{Synthetic tree evaluation: cumulative pseudo-regret (bootstrap mean and confidence
		intervals over runs). The subplot grid sweeps tree depth $d$ (rows) and branching $k$ (columns).
		Within the grid, the left half uses an uninformed prior ($\tau=\infty$) and the right half uses
		an informative prior ($\tau=0.1$).}
	\label{fig:synth-regret}
\end{figure*}

\paragraph{Setup}
Following \citet{damPowerMeanEstimation2024}, we construct fully generated synthetic $k$-ary trees of
varying depth $d$ with \emph{dense} rewards (non-zero rewards may occur at intermediate nodes, not
only at leaves). Edge rewards are normalized to $[0,1]$, and leaf rewards may have additive Gaussian
noise with standard deviation $\sigma$. Each method runs $B=1000$ simulations from the root, and we
evaluate cumulative regret (Figure~\ref{fig:synth-regret}) and root value error
(Appendix~\ref{app:synth-extra}). For each $(k,d)$ tile we aggregate $20$ independently sampled trees
with $5$ independent runs per tree.

\paragraph{Selectors and priors}
We include \textit{PUCT} (the canonical prior-based selector in AlphaZero-style pipelines) and
\textit{UCT-P} as prior-based, variance-unaware comparators. To probe the \emph{UCT form} in the
absence of prior guidance, we include \textit{SP-UCT} (Stochastic Power UCT)
\citep{damPowerMeanEstimation2024} as a strong prior-free baseline. To contrast empirical variance
tracking with alternative uncertainty heuristics, we include \textit{Wasserstein-TS} \citep{pmlr-v267-dam25c}, which uses
Thompson sampling with a fixed prior standard deviation as its exploration scale. Finally, to test
an alternative to UCT-style selectors, we include the entropy-based \textit{RENTS}, which makes use of prior information
through initialization rather than through a term in the selector. We use the default selector
hyperparameters summarized in Appendix~\ref{app:selector-hparams}.

We provide a prior at each node derived from the tree's optimal action values (computed by dynamic
programming on the fully generated tree), and control its quality via a softmax temperature $\tau$:
$\tau=\infty$ yields a uniform prior, while smaller $\tau$ yields a more informative and sharper
prior.

\paragraph{Observations.}
With an uninformed prior ($\tau=\infty$), the variance-aware variants are consistently better across
tree depth and breadth: \textit{PUCT-V} yields lower regret than \textit{PUCT}, and \textit{UCT-V-P}
yields lower regret than \textit{UCT-P}, with the advantage growing as trees become deeper and
wider. At the same time, these prior-based selectors remain competitive overall, supporting the
claim that the prior-based form remains effective even when the prior carries no information. In
contrast, \textit{Wasserstein-TS} is competitive in smaller trees but tends to lose its advantage as
depth and branching grow, consistent with using a fixed exploration scale rather than empirical
variance tracking. With an informative prior ($\tau=0.1$), all prior-based UCT-style selectors
benefit substantially, and \textit{PUCT-V}/\textit{UCT-V-P} retain an advantage over
\textit{PUCT}/\textit{UCT-P} on top of these prior gains. \textit{RENTS} shows smaller gains,
consistent with its prior being used primarily for initialization. Again, this highlights the value
of selection-time priors, which may be derived via \emph{Inverse-RPO} for other UCT-style selectors
as well.

\subsection{MinAtar Evaluation}

\begin{figure*}[!t]
	\centering
	\includegraphics[width=0.48\linewidth]{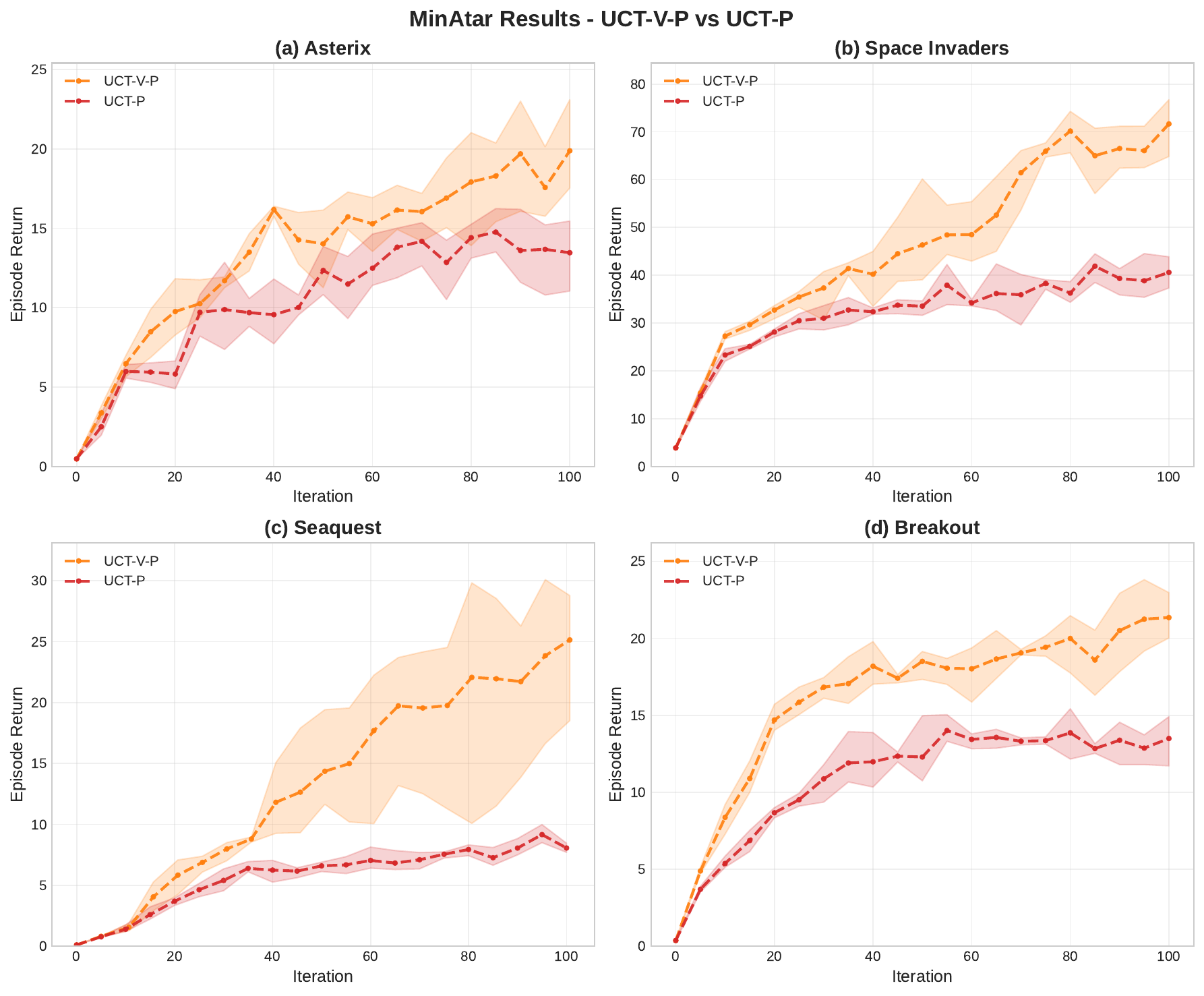}
	\hfill
	\includegraphics[width=0.48\linewidth]{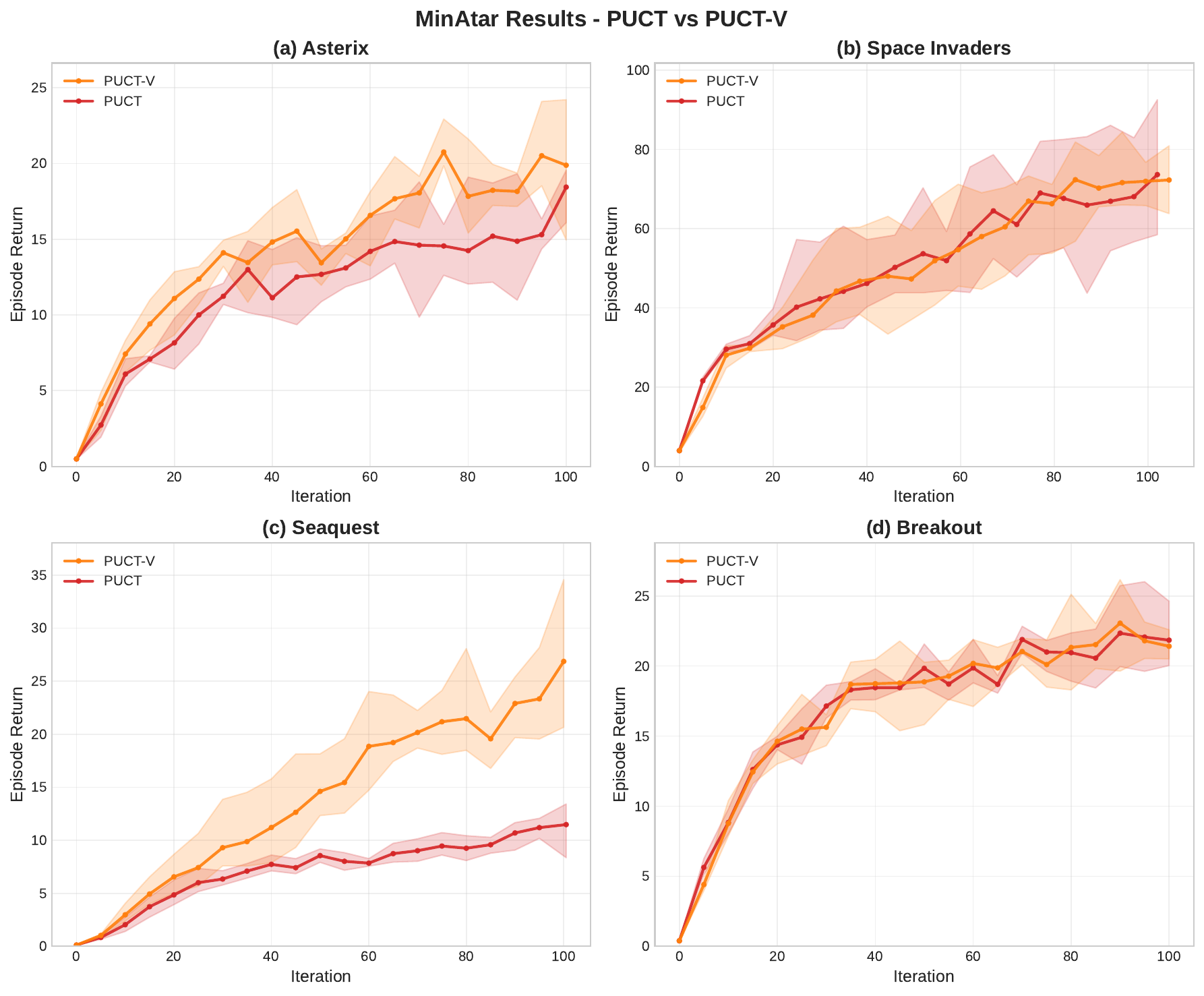}
	\caption{Average returns on the \texttt{MinAtar} suite with $N_{\mathrm{sim}}=64$. Evaluation is
		performed in batches of $256$ per seed (at least $3$ seeds), using only the trained policy head
		without MCTS. \textbf{Left:} \textit{UCT-V-P} vs.\ \textit{UCT-P}. \textbf{Right:} \textit{PUCT}
		vs.\ \textit{PUCT-V}. Solid lines indicate mean returns, and shaded regions show the
		corresponding best--worst range across seeds.}
	\label{fig:minatar-comparisons}
\end{figure*}

To evaluate in a challenging setting, we use the \texttt{MinAtar} suite
\citep{youngMinAtarAtariInspiredTestbed2019}, a widely referenced benchmark offering stochastic and
deterministic Atari-style environments that preserve the core dynamics of the original games while
being computationally efficient\footnote{We exclude the \textit{freeway} environment, as all
	evaluated algorithms consistently fail to achieve learning progress there.}. We access
\texttt{MinAtar} through the \texttt{PGX} interface
\citep{koyamadaPgxHardwareAcceleratedParallel2023}, which provides JAX-compatible environments and
an open-source AlphaZero training script that we adapt for our experiments. The search/training
pipeline is kept fixed across selectors to ensure a controlled comparison.

Unless otherwise noted, we run $N_{\mathrm{sim}}=64$ simulations per move to generate training data.
Evaluation is conducted at regular intervals in batches of $256$ trajectories per seed, and with at
least three seeds the per-checkpoint estimates are sufficiently stable for meaningful comparisons.
We adopt the network and optimization settings summarized in Table~\ref{tab:hparams}, holding
hyperparameters constant across all methods to isolate the effect of the selection rule.  Finally,
we evaluate the learned \emph{policy head} without search to assess representation and policy
quality directly and to avoid confounding from test-time MCTS.

\textbf{Observations}
Empirically, the measured wall-clock time per training step and per evaluation is essentially
identical across selectors, indicating that the proposed variance-aware MCTS and selection rules
incur no additional compute overhead. Figure~\ref{fig:minatar-comparisons} reports the average
return of the trained policy head under all benchmarked selection rules.  We compare
\textit{UCT-V-P} to \textit{UCT-P} (heuristic-free) and \textit{PUCT-V} to \textit{PUCT}
(heuristic-based).  Across all environments, the variance-aware selectors match or exceed their
variance-unaware baselines.  In particular, \textit{UCT-V-P} consistently outperforms
\textit{UCT-P}, showing that variance adjustment alone can substantially improve exploration.  For
the heuristic-based variants, \textit{PUCT-V} surpasses \textit{PUCT} on the stochastic games
\textit{Asterix} and \textit{Seaquest}, and performs comparably on deterministic ones.

Appendix~\ref{app:animal-shogi} provides an additional evaluation on Animal Shogi, a deterministic
two-player board game \citep{koyamadaPgxHardwareAcceleratedParallel2023}. In this sparse-reward
setting, the variance-aware selectors \textit{UCT-V-P} and \textit{PUCT-V} again give higher returns
than their variance-unaware counterparts. Overall, the results indicate that incorporating variance
estimates into prior-based MCTS selection can improve performance without measurable computational
overhead and with only minor modifications to the selection rule.

\section{Related Work}
\label{sec:related-work}

\paragraph{AlphaZero family and prior-based tree policies.}
Planning with MCTS coupled to learned function approximators became prominent with AlphaGo
\citep{silverMasteringGameGo2016} and was iterated upon by AlphaZero
\citep{silverGeneralReinforcementLearning2018} and MuZero \citep{schrittwieserMasteringAtariGo2020}.
Furthermore, Stochastic MuZero \citep{antonoglouPLANNINGSTOCHASTICENVIRONMENTS2022} handles
stochastic dynamics while retaining \textit{PUCT}, whereas Gumbel MuZero
\citep{danihelkaPOLICYIMPROVEMENTPLANNING2022} adopts a Gumbel-based policy-improvement objective
explicitly cast as regularized policy optimization (RPO).  A unifying ingredient in these systems is
a \emph{prior-based} tree policy that injects a policy prior into the exploration bonus.
Empirically, \textit{PUCT} (and close relatives) has become the de facto choice across domains
\citep{kemmerlingGamesSystematicReview2024}.

\paragraph{UCT family and stronger UCB bonuses.}
Beyond \textit{UCB1}, theoretically grounded UCT variants continue to be proposed
\citep{browneSurveyMonteCarlo2012}. Among such developments, variance-aware Bernstein bonuses
offer tighter instance-dependent guarantees under bounded rewards, which is why we
select \textit{UCB-V} \citep{audibertExplorationExploitationTradeoff2009} as our base.  Recent work
explores alternative distributional assumptions (e.g., Gaussian and extreme-value regimes) with
tailored regret analyses for classical planning
\citep{wissowScaleAdaptiveBalancingExploration2024,asaiExtremeValueMonte2024}.  Notably, these
methods are not prior-based by construction, so systematic prior-based extensions remain largely
missing in the literature.

\paragraph{Bayesian MCTS.}
Variance-aware and uncertainty-quantifying approaches to MCTS are active research directions.
Bayesian variants (\textit{Bayes-UCT1/2}) maintain posteriors over node values and act via
uncertainty bands \citep{tesauroBayesianInferenceMonteCarlo2012}; recent work explores richer
uncertainty models and online inference
\citep{greshlerBayesianApproachOnline2024,chenBayesAdaptiveMonte2025}. While compelling, these
methods typically introduce additional modelling choices, extra hyperparameters, and nontrivial
bookkeeping. Our proposed variance-aware prior-based tree policies based on \textit{UCB-V} likewise
bring (frequentist) uncertainty quantification into the selection rule, yet integrate as drop-in
replacements in the widely adopted AlphaZero-style MCTS with minimal changes.

\paragraph{Regularized policy optimization (RPO) and MCTS.}
Regularization-based views of RL connect policy improvement to convex programs with divergence
penalties \citep{neuUnifiedViewEntropyregularized2017,geistTheoryRegularizedMarkov2019}.
\citet{grillMonteCarloTreeSearch2020} brought this perspective to MCTS, thereby providing a
retrospective theoretical understanding for prior-based tree policies such as \textit{PUCT}.
Follow-up analyses developed regret bounds for RPO-guided MCTS and studied entropy-based
regularizers and backup operators \citep{damConvexRegularizationMonteCarlo2021}. Complementing
entropy-centric analyses, we focus on UCT-style bonuses by deriving variance-aware, prior-based
selectors with matching RPO objectives (Eqs.~\ref{eq:L-uct-v-p} and~\ref{eq:L-puct-v}).

\section{Conclusion and Future Work}
\label{sec:conclusion}

In this paper, we (1) proposed \emph{Inverse-RPO}, a principled framework to derive prior-based UCTs
from their prior-free base forms, and (2) instantiated this framework by deriving two prior-based
versions of \textit{UCB-V}.  The resulting variance-aware prior-based tree-policies,
\textit{UCT-V-P} and \textit{PUCT-V}, leverage variance estimates to improve search efficiency and
outperform existing prior-based tree-policies \textit{UCT-P} and \textit{PUCT} across multiple
benchmarks, with minimal implementation overhead.

Beyond the empirical results, our derivations of \textit{UCT-V-P} and \textit{PUCT-V} via the
\emph{Inverse-RPO} pipeline yield two RPO objectives that can be used as policy-training
targets when casting MCTS as an optimization problem in future work. Another avenue for future work
is to augment the network with a learned variance head, placed alongside the standard value and
policy heads in the AlphaZero family, to refine search-based variance estimates and further improve
the stability and performance of variance-aware prior-based UCTs. Finally, we invite the community
to revisit the well-grounded UCB literature through this lens and make principled use of its depth
by systematically deriving yet underexplored prior-based UCTs.


\bibliographystyle{unsrtnat}
\bibliography{references}

\newpage

\clearpage
\appendix
\onecolumn
\raggedbottom
\aistatstitle{Variance-Aware Prior-Based Tree Policies for Monte Carlo Tree Search: \\
	Supplementary Materials}

\section{Supplementary Code Release}
\label{app:code}

This section points readers to the released implementation and highlights where the variance-aware MCTS and
tree-policy changes live.

The full source code and reproduction instructions are available at:
\href{https://github.com/Max-We/inverse-rpo}{github.com/Max-We/inverse-rpo}.

\begin{itemize}
	\item \textbf{Modifications to \texttt{mctx}.} This includes the variance-aware extensions of the MCTS
	      backpropagation routine, together with the implementation of the proposed variance-aware tree
	      policies \textit{UCT-V-P} and \textit{PUCT-V}. These modifications are fully integrated into the
	      existing \texttt{mctx} API and intended as minimal drop-in changes.

	\item \textbf{Training and Evaluation.} We include the \texttt{pgx} environments
	      \citep{koyamadaPgxHardwareAcceleratedParallel2023} with a training script adapted for the
	      \texttt{MinAtar} experiments. This script reproduces all experimental results presented in
	      Section~\ref{sec:experiments}.
\end{itemize}

\section{Monte Carlo Tree Search: Four Stages}
\label{app:mcts}

For completeness, we briefly recall the four canonical stages of MCTS:

\begin{enumerate}
	\item \textbf{Selection.} Starting from the root, recursively select child nodes according to a tree
	      policy (e.g., \textit{UCT1} or \textit{PUCT}) until reaching a leaf node.
	\item \textbf{Expansion.} If the leaf node corresponds to a non-terminal state, expand the tree
	      by adding a child node to the selected leaf node. Some implementations also expand terminal
	      nodes, setting the discount-factor $\gamma$ to zero.
	\item \textbf{Evaluation (Simulation).} Evaluate the expanded node with a neural network
	      (AlphaZero approach) or by conducting rollouts under a rollout policy (classical approach).
	\item \textbf{Backpropagation.} Propagate the evaluated node statistics back through the visited path,
	      updating statistics at each parent node. These statistics typically include the information required
	      by the tree policy, such as the value and visit count of a node.
\end{enumerate}

These four steps are repeated for a fixed number of simulations, after which an action is chosen
based on the statistics of the children of the root node. In the AlphaZero family of algorithms, a
neural network is additionally trained on the root statistics and node statistics such as the value
are \textbf{normalized} to conform to the UCB assumptions.

\section{Inverse--RPO Derivations for Variance-Aware UCTs}
\label{app:uctv-derivations}

This appendix provides the complete inverse--RPO derivations that lead to the prior-based,
variance-aware selection rules and objectives stated in the main text
(\S\ref{sec:puct-v}; cf.\ \eqref{eq:uct-v-p}, \eqref{eq:puct-v},
\eqref{eq:L-uct-v-p}, \eqref{eq:L-puct-v}).
We reuse the empirical selector $\hat\pi$ from \eqref{eq:selector}.

\subsection{UCT-V-P (derivation)}
Starting from the prior-free UCB-V/UCT-V bonus, we construct an RPO objective whose marginal-gain (greedy)
rule recovers the bonus, then lift the resulting (separable) regularizers to $f$-divergence terms
with prior $\pi_\theta$ to obtain the main-text UCT-V-P objective and
selection rule.

\paragraph{1. Decompose the exploration bonus into action-independent and action-dependent factors.}
Starting from the variance-aware UCT score \eqref{eq:uct-v}, the exploration bonus decomposes into
action-independent scaling factors and action-dependent weighting functions as
\begin{equation}
	\label{eq:uct-v-factorization}
	B^{\mathrm{UCT\mbox{-}V}}(N,n_a,\hat\sigma_a^2)
	= \lambda_{N}^{\mathrm{UCT\mbox{-}V-1}}\,h_{\mathrm{H}}\!\big(\hat\pi(a),\hat\sigma_a\big)
	+ \lambda_{N}^{\mathrm{UCT\mbox{-}V-2}}\,h_{\mathrm{KL}}\!\big(\hat\pi(a)\big),
	\quad
	h_{\mathrm{H}}(x,\sigma)=\tfrac{\sigma}{\sqrt{x}},
	\quad
	h_{\mathrm{KL}}(x)=\tfrac{1}{x},
\end{equation}
with scaling terms
\begin{equation}
	\lambda_{N}^{\mathrm{UCT\mbox{-}V-1}} = c_1 \tfrac{\sqrt{\log N}}{\sqrt{|\mathcal A|+N}},
	\qquad
	\lambda_{N}^{\mathrm{UCT\mbox{-}V-2}} = c_2 \tfrac{\log N}{|\mathcal A|+N}.
\end{equation}

\paragraph{2. Define a weighted sum of (separable) regularizers.}
Choose convex functions whose derivatives match the negative action-dependent weighting functions
$h_{\mathrm{H}}$ and $h_{\mathrm{KL}}$:
\begin{equation}
	f^{\mathrm{H}}(x,\sigma)=2\,\sigma\,(1-\sqrt{x}) \quad\Rightarrow\quad {f^{\mathrm{H}}}'(x,\sigma)=-\tfrac{\sigma}{\sqrt{x}},
	\qquad
	f^{\mathrm{KL}}(x)=-\log x \quad\Rightarrow\quad {f^{\mathrm{KL}}}'(x)=-\tfrac{1}{x}.
\end{equation}
This yields the RPO with a \emph{weighted sum of (separable) regularizers generated by convex
	functions}:
\begin{equation}
	\label{eq:L-uct-v}
	L_{\mathrm{UCT\mbox{-}V}}(y)
	= \mathbf{q}^\top y
	- \lambda_{N}^{\mathrm{UCT\mbox{-}V-1}} \sum_a f^{\mathrm{H}}\!\big(y_a,\hat\sigma_a\big)
	- \lambda_{N}^{\mathrm{UCT\mbox{-}V-2}} \sum_a f^{\mathrm{KL}}\!\big(y_a\big),
\end{equation}
whose marginal-gain rule in $n_a$ recovers \eqref{eq:uct-v-factorization}.

\paragraph{3. Lift the (separable) regularizers to $f$-divergence terms.}
Lift the (separable) regularizers to $f$-divergence terms with prior $\pi_\theta$,
noting that the variance term becomes variance-weighted:
\begin{equation}
	\label{eq:uctvp-weighted-lifts}
	\begin{aligned}
		D_{\mathrm{H}}^{\hat\sigma}(\pi_\theta,y)
		 & = \sum_{a\in\mathcal A} \pi_\theta(a)\;
		2\,\hat\sigma_a\!\left(1-\sqrt{\tfrac{y_a}{\pi_\theta(a)}}\right), \\
		D_{\mathrm{KL}}(\pi_\theta,y)
		 & = \sum_{a\in\mathcal A} \pi_\theta(a)\;
		\log\!\frac{\pi_\theta(a)}{y_a}.
	\end{aligned}
\end{equation}
The resulting prior-based objective is exactly the form stated in the main text:
\begin{equation*}
	\text{(cf.\ \eqref{eq:L-uct-v-p})}\qquad
	L_{\mathrm{UCT\mbox{-}V\mbox{-}P}}(y)
	= \mathbf{q}^\top y
	- \lambda_{N}^{\mathrm{UCT\mbox{-}V-1}}\, D_{\mathrm{H}}^{\hat\sigma}(\pi_\theta,y)
	- \lambda_{N}^{\mathrm{UCT\mbox{-}V-2}}\, D_{\mathrm{KL}}(\pi_\theta,y).
\end{equation*}

\paragraph{4. Recover the prior-based UCT rule.}
Taking the directional derivative in $n_a$ yields the greedy expansion rule reported in the main text:
\begin{equation*}
	\text{(cf.\ \eqref{eq:uct-v-p})}\qquad
	S_a^{\mathrm{UCT\mbox{-}V\mbox{-}P}}(q,n,N)
	= q_a
	+ c_1 \cdot \hat\sigma_a \, \sqrt{ \pi_\theta(a)\, \tfrac{\log N}{1+n_a} }
	+ c_2 \cdot \pi_\theta(a) \, \tfrac{\log N}{1+n_a}.
\end{equation*}

\subsection{PUCT-V (heuristic variant; derivation)}
We repeat the same inverse--RPO steps for the heuristic UCT-V variant, yielding an
$f$-divergence objective and the main-text PUCT-V selector.

\paragraph{1. Decompose the exploration bonus into action-independent and action-dependent factors.}
For the heuristic variant \eqref{eq:uct-v-h}, the bonus decomposes into action-independent scaling
factors and action-dependent weighting functions as
\begin{equation}
	\label{eq:uct-v-h-factorization}
	B^{\mathrm{UCT\mbox{-}V{-}H}}(N,n_a,\hat\sigma_a^2)
	= \lambda_{N}^{\mathrm{UCT\mbox{-}V\mbox{-}H-1}}\,h_{\mathrm{H}}\!\big(\hat\pi(a),\hat\sigma_a\big)
	+ \lambda_{N}^{\mathrm{UCT\mbox{-}V\mbox{-}H-2}}\,h_{\mathrm{KL}}\!\big(\hat\pi(a)\big),
\end{equation}
with
\begin{equation}
	h_{\mathrm{H}}(x,\sigma)=\tfrac{\sigma}{x},
	\qquad
	h_{\mathrm{KL}}(x)=\tfrac{1}{x},
	\qquad
	\lambda_{N}^{\mathrm{UCT\mbox{-}V\mbox{-}H-1}} = c_1 \tfrac{\sqrt{N}}{|\mathcal A|+N},
	\qquad
	\lambda_{N}^{\mathrm{UCT\mbox{-}V\mbox{-}H-2}} = c_2 \tfrac{\log N}{|\mathcal A|+N}.
\end{equation}

\paragraph{2. Define a weighted sum of (separable) regularizers.}
Choose convex functions whose derivatives match the negative action-dependent weighting functions
$h_{\mathrm{H}}$ and $h_{\mathrm{KL}}$:
\begin{equation}
	f^{\mathrm{KL},\sigma}(x,\sigma)=-\sigma \log x
	\quad\Rightarrow\quad
	{f^{\mathrm{KL},\sigma}}'(x,\sigma)=-\tfrac{\sigma}{x},
	\qquad
	f^{\mathrm{KL}}(x)=-\log x
	\quad\Rightarrow\quad
	{f^{\mathrm{KL}}}'(x)=-\tfrac{1}{x}.
\end{equation}
This yields the RPO with a \emph{weighted sum of (separable) regularizers generated by convex
	functions}:
\begin{equation}
	\label{eq:L-uct-v-h}
	L_{\mathrm{UCT\mbox{-}V{-}H}}(y)
	= \mathbf{q}^\top y
	- \lambda_{N}^{\mathrm{UCT\mbox{-}V\mbox{-}H-1}} \sum_a f^{\mathrm{KL},\sigma}\!\big(y_a,\hat\sigma_a\big)
	- \lambda_{N}^{\mathrm{UCT\mbox{-}V\mbox{-}H-2}} \sum_a f^{\mathrm{KL}}\!\big(y_a\big).
\end{equation}
whose marginal-gain rule recovers \eqref{eq:uct-v-h-factorization}.

\paragraph{3. Lift the (separable) regularizers to $f$-divergence terms.}
Lifting to $f$-divergence terms with prior $\pi_\theta$, the variance term becomes
a variance-weighted reverse-KL $f$-divergence term:
\begin{equation}
	\label{eq:puctv-weighted-lifts}
	\begin{aligned}
		D_{\mathrm{KL}}^{\hat\sigma}(\pi_\theta,y)
		 & = \sum_{a\in\mathcal A} \pi_\theta(a)\;
		\hat\sigma_a \log\!\frac{\pi_\theta(a)}{y_a}, \\
		D_{\mathrm{KL}}(\pi_\theta,y)
		 & = \sum_{a\in\mathcal A} \pi_\theta(a)\;
		\log\!\frac{\pi_\theta(a)}{y_a}.
	\end{aligned}
\end{equation}
This yields the prior-based objective reported in the main text:
\begin{equation*}
	\text{(cf.\ \eqref{eq:L-puct-v})}\qquad
	L_{\mathrm{PUCT\mbox{-}V}}(y)
	= \mathbf{q}^\top y
	- \lambda_{N}^{\mathrm{UCT\mbox{-}V\mbox{-}H-1}}\, D_{\mathrm{KL}}^{\hat\sigma}(\pi_\theta,y)
	- \lambda_{N}^{\mathrm{UCT\mbox{-}V\mbox{-}H-2}}\, D_{\mathrm{KL}}(\pi_\theta,y).
\end{equation*}

\paragraph{4. Recover the prior-based UCT rule.}
Taking the directional derivative in $n_a$ yields the selection rule as stated:
\begin{equation*}
	\text{(cf.\ \eqref{eq:puct-v})}\qquad
	S_a^{\mathrm{PUCT\mbox{-}V}}(q,n,N)
	= q_a
	+ c_1 \cdot \pi_\theta(a)\, \hat\sigma_a \, \tfrac{\sqrt{N}}{1+n_a}
	+ c_2 \cdot \pi_\theta(a) \, \tfrac{\log N}{1+n_a}.
\end{equation*}

\section{Selector Hyperparameters (Bandit and Synthetic Tree)}
\label{app:selector-hparams}

\begin{table}[h]
	\centering
	\caption{Default selector hyperparameters used in the bandit and synthetic-tree experiments (unless stated otherwise).}
	\label{tab:selector-hparams}
	\small
	\begingroup
	\setlength{\tabcolsep}{5pt}
	\begin{tabular}{l l r}
		\hline
		\textbf{Symbol}           & \textbf{Used in}                                          & \textbf{Default}     \\
		\hline
		$c$                       & \textit{UCT1}/\textit{UCT}, \textit{UCT-P}                & $\sqrt{2}$           \\
		$c_{\mathrm{PUCT}}$       & \textit{PUCT}                                             & $1.25$               \\
		$c_1$                     & \textit{UCT-V-P}, \textit{PUCT-V}                         & $\sqrt{2}$           \\
		$c_2$                     & \textit{UCT-V-P}, \textit{PUCT-V}                         & $3$                  \\
		$\alpha_{\mathrm{var}}$   & \textit{UCT-V-P}, \textit{PUCT-V} variance initialization & $10^{-3} \cdot 0.25$ \\
		$\tau_{\mathrm{E3W}}$     & \textit{RENTS}                                            & $10^{-3}$            \\
		$\epsilon_{\mathrm{E3W}}$ & \textit{RENTS}                                            & $1.0$                \\
		$c_{\mathrm{SP}}$         & \textit{SP-UCT}                                           & $0.25$               \\
		$c_{\mathrm{Exp1}}$       & \textit{SP-UCT}                                           & $0.25$               \\
		$c_{\mathrm{Exp2}}$       & \textit{SP-UCT}                                           & $0.5$                \\
		$p$                       & \textit{SP-UCT} power                                     & $2.0$                \\
		$c_{\mathrm{W}}$          & Wasserstein-TS                                            & $1.0$                \\
		$\sigma_{\mathrm{prior}}$ & Wasserstein-TS                                            & $30.0$               \\
		\hline
	\end{tabular}
	\endgroup
\end{table}

\section{Additional Synthetic Tree Results}
\label{app:synth-extra}

\begin{figure}[H]
	\centering
	\includegraphics[width=0.95\linewidth]{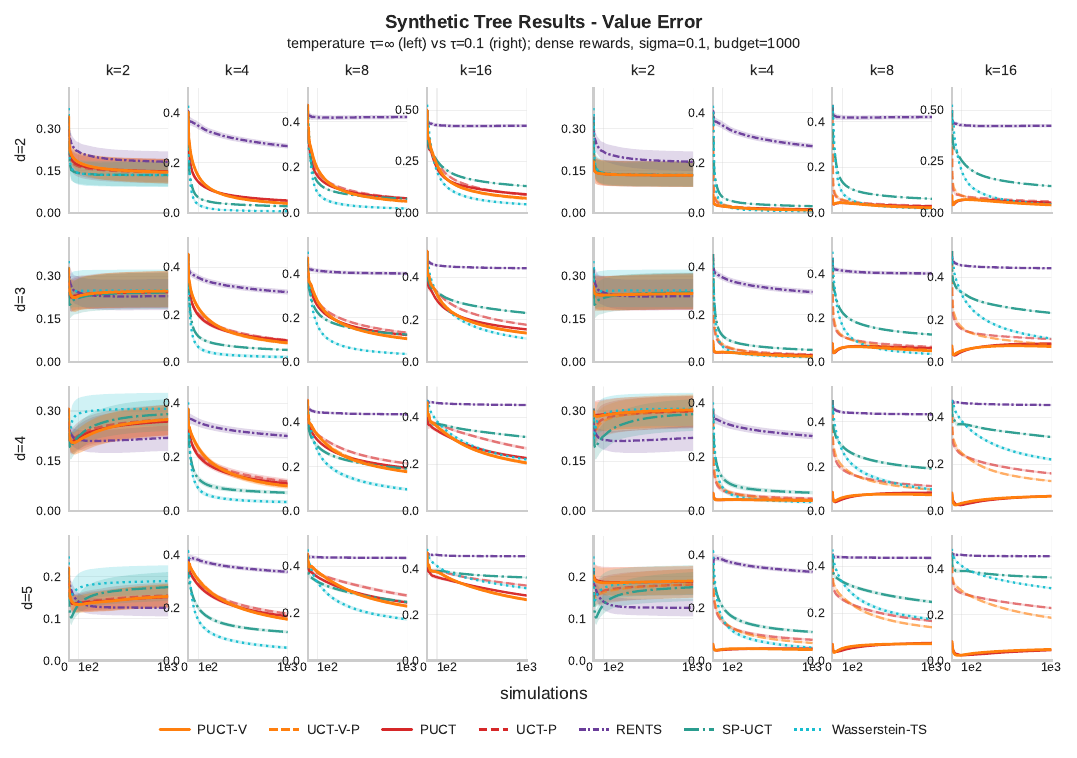}
	\caption{Synthetic tree evaluation: root value-error (bootstrap mean and confidence intervals over
		runs). The subplot grid sweeps tree depth $d$ (rows) and branching $k$ (columns). Within the
		grid, the left half uses an uninformed prior ($\tau=\infty$) and the right half uses an
		informative prior ($\tau=0.1$).}
	\label{fig:synth-value-error}
\end{figure}

\section{Additional Animal Shogi Results}
\label{app:animal-shogi}

\begin{figure}[H]
	\centering
	\includegraphics[width=0.48\linewidth]{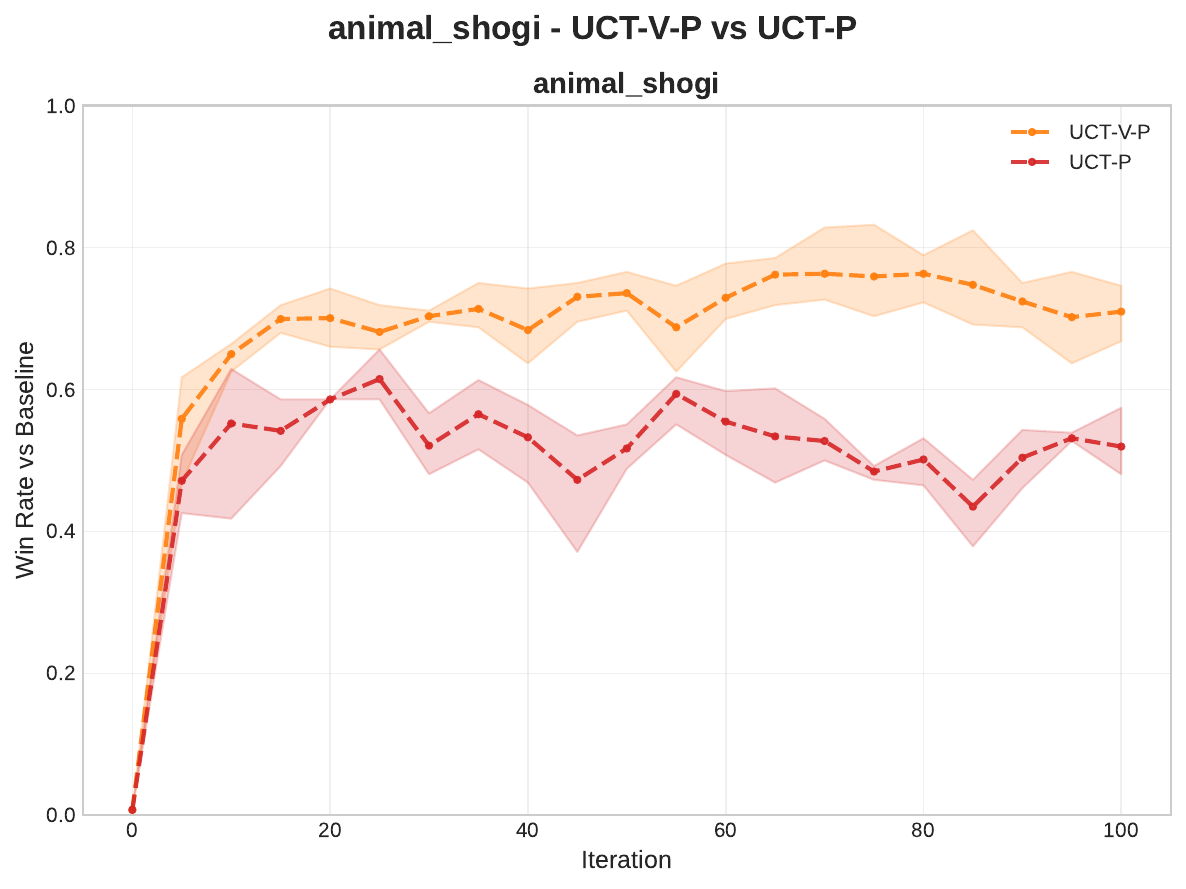}
	\hfill
	\includegraphics[width=0.48\linewidth]{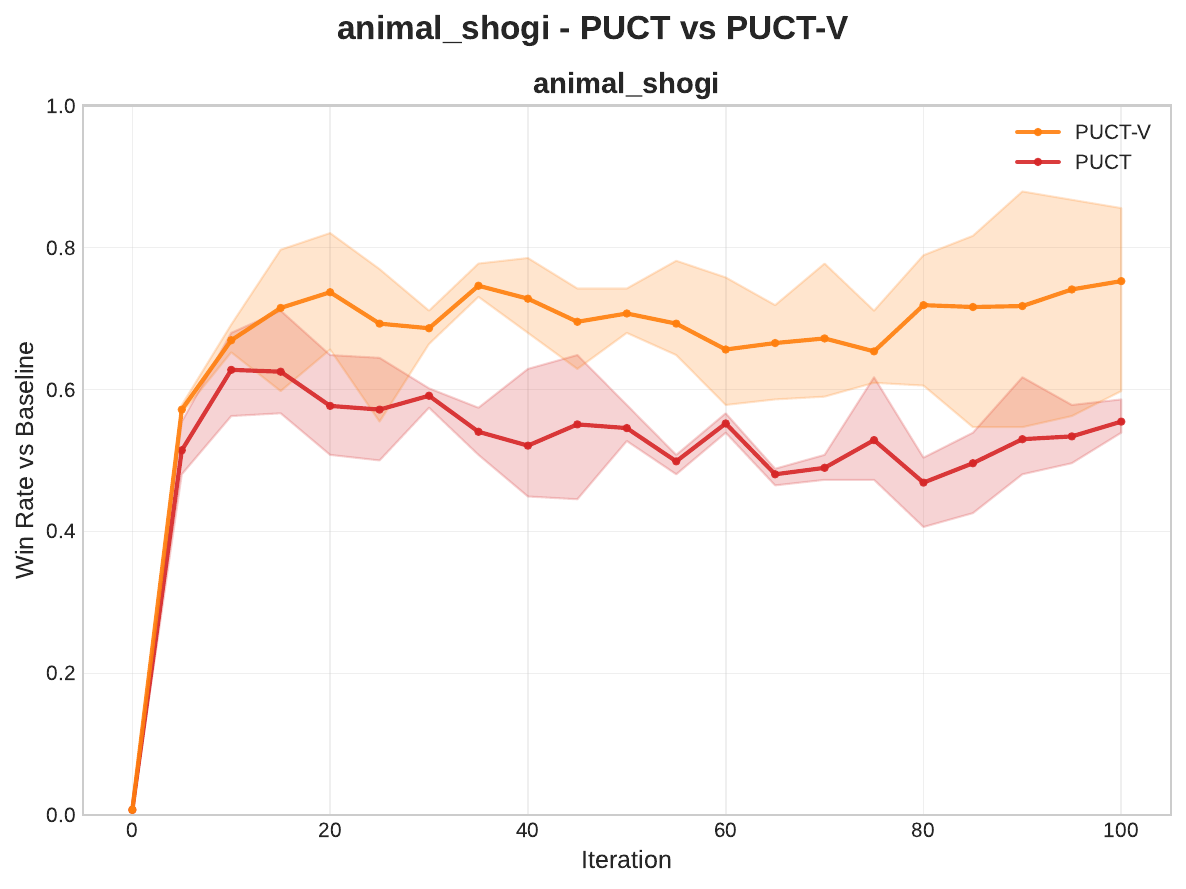}
	\caption{Win rate against the \texttt{PGX} Animal Shogi baseline with $N_{\mathrm{sim}}=64$.
		Evaluation is performed every 5 training iterations with $128$ games per seed (3 seeds), using
		$100$ training iterations in total. \textbf{Left:} \textit{UCT-V-P} vs.\ \textit{UCT-P}.
		\textbf{Right:} \textit{PUCT} vs.\ \textit{PUCT-V}. Solid lines indicate mean win rates, and
		shaded regions show the corresponding best--worst range across seeds.}
	\label{fig:animal-shogi-comparisons}
\end{figure}

At the final training iteration, the variance-aware selectors achieve higher win rates against the \texttt{PGX} baseline: \textit{PUCT-V} reaches $75\%$ mean win rate across seeds (range: $60\%$--$86\%$) versus $55\%$ for \textit{PUCT} (range: $54\%$--$59\%$), and \textit{UCT-V-P} reaches $71\%$ (range: $67\%$--$75\%$) versus $52\%$ for \textit{UCT-P} (range: $48\%$--$57\%$).

\section{Hyperparameters for the \texttt{MinAtar} Experiments}
\label{app:hparams}

In our experiments we used the hyperparameters in Table~\ref{tab:hparams} consistently across all benchmarks.

\begin{table}[h]
	\centering
	\caption{Key hyperparameters used for \texttt{MinAtar} experiments.}
	\label{tab:hparams}
	\begin{tabular}{l l}
		\hline
		\textbf{Hyperparameters}         & \textbf{Value}                 \\
		\hline
		Iterations                       & 100                            \\
		Simulations ($N_{\mathrm{sim}}$) & 64                             \\
		Self-play batch size             & 256                            \\
		Max.\ steps per episode          & 256                            \\
		Training batch size              & 1024                           \\
		Learning rate                    & $1\times 10^{-3}$              \\
		Discount factor $\gamma$         & 0.99                           \\
		Evaluation interval              & every 5 iterations             \\
		Network architecture             & 6-layer ResNet-v2, 32 channels \\
		\hline
	\end{tabular}
\end{table}

\end{document}